\documentclass{article}

\usepackage{microtype}
\usepackage{graphicx}
\usepackage{subcaption}
\usepackage{booktabs}
\usepackage{hyperref}
\usepackage{multirow}
\usepackage{xcolor}
\usepackage{tikz}
\usetikzlibrary{positioning, arrows.meta, calc, fit, backgrounds, shapes.geometric, decorations.pathreplacing}
\usepackage{caption}

\usepackage[preprint]{icml2026}

\usepackage{amsmath}
\usepackage{amssymb}
\usepackage{mathtools}
\usepackage{amsthm}

\usepackage[capitalize,noabbrev]{cleveref}

\theoremstyle{plain}

\DeclareMathOperator*{\argmax}{arg\,max}

\newcommand{\std}[1]{{\scriptsize$\pm$#1}}
\newcommand{\chimera}{\textsc{Chimera-Bench}}
\newcommand{\evostruct}{\textsc{EvoStruct}}
\usepackage[normalem]{ulem}
\newcommand{\second}[1]{\uline{#1}}

\begin{document}

\twocolumn[
  \icmltitle{EvoStruct: Bridging Evolutionary and Structural Priors for Antibody CDR Design via Protein Language Model Adaptation}

  \icmlsetsymbol{equal}{*}

  \begin{icmlauthorlist}
    \icmlauthor{Mansoor Ahmed}{gsu,gt}
    \icmlauthor{Sujin Lee}{gt}
    \icmlauthor{Umar Khayaz}{gt}
    \icmlauthor{Murray Patterson}{gsu}
  \end{icmlauthorlist}

    \icmlaffiliation{gsu}{Georgia State University, Atlanta, USA}
    \icmlaffiliation{gt}{Georgia Institute of Technology, Atlanta, USA}
    \icmlcorrespondingauthor{Mansoor Ahmed}{mahmed76@student.gsu.edu}

  \icmlkeywords{antibody design, CDR, protein language model, graph neural network, equivariant, adapter}

  \vskip 0.3in
]

\printAffiliationsAndNotice{}

\begin{abstract}
Equivariant graph neural network (GNN) methods for antibody complementarity-determining region (CDR) design achieve the highest sequence recovery but suffer from severe vocabulary collapse. The current best GNN methods over-predict very few amino acids, such as tyrosine and glycine, while ignoring functionally important residues. We trace this failure to GNN encoders learning amino acid distributions \emph{de novo} from limited structural data, discarding substitution patterns encoded in evolutionary databases. To resolve this, we propose \evostruct{}, which bridges a frozen protein language model (PLM) with 3D structural context from an E(3)-equivariant GNN via a cross-attention adapter. Unlike prior PLM-structure adapters for general protein design, \evostruct{} targets the vocabulary collapse problem specific to CDR design through progressive PLM unfreezing and R-Drop consistency regularization. On the \chimera{} benchmark, \evostruct{} achieves the highest amino acid recovery and lowest perplexity among several antibody design methods, improving sequence recovery by 16\% and reducing perplexity by 43\% relative to the best GNN baselines, while recovering 2.3$\times$ greater amino acid diversity and the highest binding-pair correlation with ground truth.
\end{abstract}

\section{Introduction}
\label{sec:intro}

Antibodies bind antigens through their complementarity-determining regions (CDRs), six hypervariable loops whose sequence and structure determine binding specificity~\citep{chothia1987canonical}. Computational CDR design methods condition on antigen structure to generate sequences and backbone conformations for these loops~\citep{luo2022diffab,kong2022mean,kong2023dymean,wu2025raad}. A growing body of evidence shows that existing methods largely fail to leverage antigen information. Predictions remain nearly unchanged when the antigen is removed~\citep{mason2025inversefoldingbenchmark}, and BLOSUM substitution matrices explain model outputs as well as learned likelihoods~\citep{uccar2025blosum,chinery2024simple}.

A systematic evaluation on the \chimera{} benchmark~\citep{ahmed2026chimerabench} reveals that GNN methods with greedy decoding exhibit three linked failure modes. \textbf{Vocabulary collapse} reduces predicted amino acids to a fraction of the native diversity, with models overpredicting tyrosine and glycine while ignoring functionally important residues. This echoes the germline bias identified by \citet{olsen2024germline} in antibody language models, but manifests more severely in structure-conditioned GNN methods. \textbf{Poor binding-pair learning} limits paratope-epitope amino acid pair correlation, indicating that no baseline captures the full binding-pair structure. \textbf{Contact position weakness} produces dramatically lower recovery at antigen-contacting positions compared to non-contacting positions, confirming that models learn general CDR statistics rather than antigen-specific binding preferences.

These failures share a common root. GNN encoders learn amino acid distributions \emph{de novo} from limited structural training data (${\sim}$3,000 complexes), discarding the substitution patterns encoded in evolutionary sequence databases spanning hundreds of millions of proteins. Recent work on structure-informed PLM adaptation~\citep{zheng2023lmdesign} has shown that bridging frozen protein language models with structural encoders via cross-attention adapters preserves the PLM's vocabulary calibration for general protein design. We apply this paradigm to CDR design with \evostruct{}, which bridges a frozen ESM-2~\citep{lin2023evolutionary} with an E(3)-equivariant GNN encoder through a cross-attention adapter that operates in ESM-2's representation space. Unlike general inverse folding adapters~\citep{zheng2023lmdesign,shanehsazzadeh2023igdesign}, \evostruct{} targets the vocabulary collapse problem specific to CDR design through progressive PLM unfreezing~\citep{howard2018universal} and R-Drop consistency regularization~\citep{liang2021rdrop}, and provides a systematic failure mode analysis quantifying the diversity-accuracy tradeoff that constrains existing GNN methods.

Our contributions are:
\begin{enumerate}
    \item A quantitative diagnosis of vocabulary collapse in CDR design using effective vocabulary ($V_\text{eff}$) and binding-pair correlation metrics, revealing that GNN methods recover only 20--35\% of native amino acid diversity.
    \item An adaptation of PLM-structure bridging for CDR design with progressive unfreezing and R-Drop regularization.
    \item Empirical evidence that PLM vocabulary calibration transfers to CDR design, breaking the diversity-accuracy tradeoff with 16\% higher sequence recovery, 43\% lower perplexity, and 2.3$\times$ greater amino acid diversity than the best GNN baselines.
\end{enumerate}

The rest of this paper is organized as follows. \Cref{sec:related} discusses related work, and \Cref{sec:preliminaries} defines the task, graph construction, and failure mode analysis. \Cref{sec:method} describes the \evostruct{} architecture. \Cref{sec:experiments} presents experimental results and failure mode resolution. \Cref{sec:conclusion} concludes.

\section{Related Work}
\label{sec:related}

\paragraph{GNN-based CDR design.}
Equivariant GNN methods including MEAN~\citep{kong2022mean}, dyMEAN~\citep{kong2023dymean}, and RAAD~\citep{wu2025raad} condition on antigen through spatial message passing and achieve the highest sequence recovery among all paradigms. However, these methods predict amino acids through simple linear or MLP heads operating in the GNN's learned embedding space, which suffers from severe vocabulary collapse under greedy decoding. Multiple studies document this and broader conditioning failures in CDR design methods: \citet{mason2025inversefoldingbenchmark} show that predictions change minimally when the antigen is removed, \citet{uccar2025blosum} demonstrate that BLOSUM matrices explain model outputs as well as learned likelihoods, and \citet{chinery2024simple} find that simple computational methods can match deep learning approaches for generating diverse binder-enriched libraries. \citet{olsen2024germline} identify a related germline bias in antibody language models where overprediction of germline residues limits design utility. We provide a systematic quantification of this vocabulary collapse in structure-conditioned CDR design and demonstrate that PLM adaptation is an effective solution.

\paragraph{Generative CDR design.}
Diffusion-based methods such as DiffAb~\citep{luo2022diffab}, AbFlowNet~\citep{abir2025abflownet}, and AbMEGD~\citep{chen2025AbMEGD} model CDR generation on SE(3), while AbODE~\citep{verma2023abode} uses conjoined ODEs. RefineGNN~\citep{jin2021refinegnn} generates CDRs autoregressively without antigen input. These sampling-based methods maintain higher vocabulary diversity but achieve substantially lower sequence recovery. FlowDesign~\citep{wu2025flowdesign} addresses prior distribution quality for CDR design, targeting structural diversity rather than the sequence vocabulary collapse we identify. Among PLM-based approaches, AntiFold~\citep{hoie2025antifold} fine-tunes ESM-IF1 for antibody inverse folding with CDR-weighted masking. AbEgDiffuser~\citep{zhu2025abegdiffuser} and RADAb~\citep{wang2024radab} incorporate frozen ESM-2 features through additive injection or concatenation with GNN residue embeddings, but neither uses cross-attention adaptation or operates the sequence head in PLM embedding space.

\paragraph{Structure-informed PLM adaptation.}
LM-Design~\citep{zheng2023lmdesign} introduced the paradigm of inserting cross-attention adapters into frozen PLMs to inject structural context from GNN encoders for protein inverse folding. IgDesign~\citep{shanehsazzadeh2023igdesign} applies a similar bottleneck adapter between IgMPNN and ESM2-3B specifically for antibody design, with in vitro experimental validation. Our work builds on this paradigm but differs in motivation and training strategy. Where LM-Design and IgDesign target general sequence recovery, we diagnose and address the vocabulary collapse problem specific to CDR design. We additionally employ progressive PLM unfreezing following the schedule of \citet{howard2018universal} and R-Drop consistency regularization~\citep{liang2021rdrop}, which to our knowledge has not been applied in protein design. The shadow paratope loss from dyMEAN~\citep{kong2023dymean} provides pairwise distance supervision for CDR-epitope geometry.

\section{Preliminaries}
\label{sec:preliminaries}

\subsection{Task Definition}
\label{sec:task}

Following \chimera{}~\citep{ahmed2026chimerabench}, we formulate the CDR design task as follows. Given an antigen structure $A = \{(s_j, \mathbf{x}_j) \mid j \in V_A\}$, an epitope specification $E \subseteq V_A$, and an antibody framework $F = \{(s_i, \mathbf{x}_i) \mid i \in V_\text{FR}\}$, we design CDR residues $R = \{(s_k, \mathbf{x}_k) \mid k \in V_\text{CDR}\}$ that maximize the conditional likelihood subject to epitope contact constraints:
\begin{equation}
    R^* = \argmax_{R} \; p_\theta\!\bigl(R \mid A, E, F\bigr), \quad
    \text{s.t.} \;\; \mathcal{C}(R, A) \neq \emptyset
    \label{eq:task}
\end{equation}
where each residue has amino acid type $s_k \in \{1, \ldots, 20\}$ and C$\alpha$ coordinate $\mathbf{x}_k \in \mathbb{R}^3$. The contact set $\mathcal{C}(R, A) = \{j \in V_A \mid \exists\, k \in V_\text{CDR}\!: \|\mathbf{x}_k - \mathbf{x}_j\| < d_c\}$ denotes antigen residues contacted within cutoff $d_c$. We focus on CDR-H3, the most variable loop and primary determinant of antigen specificity~\citep{chothia1987canonical}.

\subsection{Graph Construction}
\label{sec:graph}

We represent the antibody-antigen complex as a heterogeneous graph $\mathcal{G} = (V, \mathcal{E})$. The node set $V = V_\text{HC} \cup V_\text{LC} \cup V_A \cup V_\text{glob}$ contains residue nodes from the heavy chain, light chain, and antigen, plus three global delimiter tokens ($V_\text{glob} = \{\text{BOH}, \text{BOL}, \text{BOA}\}$). Each residue node $i$ carries amino acid type $s_i \in \{1, \ldots, 20\}$ and four backbone atom coordinates $\mathbf{X}_i = [\mathbf{x}_i^\text{N}, \mathbf{x}_i^{\text{C}\alpha}, \mathbf{x}_i^\text{C}, \mathbf{x}_i^\text{O}] \in \mathbb{R}^{4 \times 3}$.

We partition the edge set $\mathcal{E}$ into 8 typed subsets that capture different structural relationships. Within each chain, we construct \emph{radial edges} connecting all pairs within a C$\alpha$ distance threshold, \emph{sequential edges} linking residues separated by one or two positions in primary sequence, and \emph{KNN edges} connecting each residue to its $K$ nearest spatial neighbors. Across chains, we add \emph{inter-chain radial edges} and \emph{inter-chain KNN edges} that enable direct communication between antibody and antigen residues. Global delimiter tokens connect to all residues in their respective chains via \emph{global-to-chain edges}. Each edge $(i, j)$ of type $t$ carries a feature vector $\mathbf{e}_{ij} \in \mathbb{R}^{d_e}$ encoding edge type, relative position in local coordinate frames, pairwise distance RBFs across four backbone atom pairs, a quaternion encoding of relative backbone orientation, and local frame direction features.

\subsection{Failure Mode Analysis}
\label{sec:failure}

We evaluate several CDR-H3 design methods on the \chimera{} benchmark and identify three systematic failure modes that motivate our approach.

\paragraph{Vocabulary Collapse.}
We define the effective vocabulary as $V_\text{eff} = \exp\!\left(-\sum_a p(a) \log p(a)\right)$, the exponentiated Shannon entropy of the predicted amino acid distribution. Native CDR-H3 sequences exhibit $V_\text{eff} \approx 15.5$, indicating near-uniform usage of all 20 amino acids. GNN methods with greedy decoding collapse to $V_\text{eff}$ of 3.0--5.5. RAAD~\citep{wu2025raad} overpredicts tyrosine by 34\% and glycine by 21\%, while rare but biochemically critical residues (tryptophan, cysteine, methionine) appear at near-zero frequency (\cref{fig:vocab_collapse}). Sampling-based methods approach native diversity ($V_\text{eff}$ 11.7--14.9) but at the cost of much lower accuracy.

\paragraph{Binding-Pair Failure.}
Binding specificity requires that the model learn which paratope amino acids pair with which epitope amino acids. We compute the correlation between predicted and ground-truth paratope-epitope amino acid pair frequencies. No baseline exceeds $r = 0.69$ (\cref{fig:binding_pairs}), and DiffAb~\citep{luo2022diffab} shows anti-correlation ($r = -0.29$), systematically predicting the wrong amino acids at interface positions. This indicates that existing conditioning mechanisms route negligible antigen information to the sequence prediction head.

\paragraph{Contact Position Weakness.}
All methods show dramatically lower amino acid recovery at antigen-contacting positions (8--23\%) compared to non-contacting positions (23--51\%). Contact positions are precisely where amino acid identity is most constrained by the binding partner, yet they are where models struggle to perform well. The gap reveals that models learn general CDR positional statistics rather than antigen-specific binding physics, and that the antigen input contributes little to prediction quality even at the positions where it matters most.

\section{Method}
\label{sec:method}

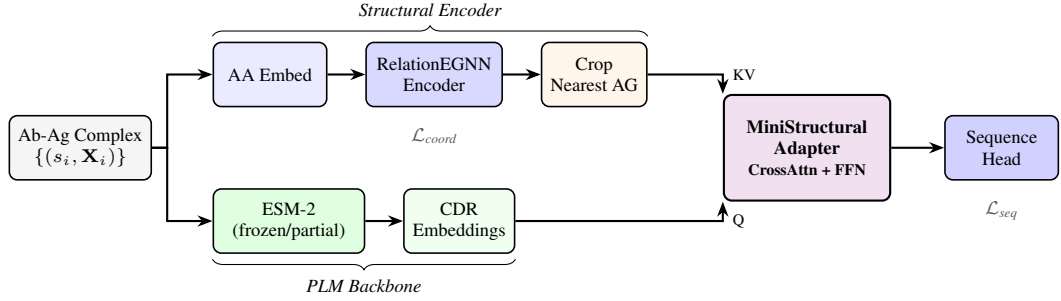
\begin{figure*}[t]
    \centering
    \begin{tikzpicture}[
        node distance=0.5cm and 0.4cm,
        block/.style={rectangle, draw, rounded corners=3pt, minimum height=0.85cm, minimum width=1.5cm, align=center, font=\scriptsize, line width=0.6pt},
        stage/.style={rectangle, draw, rounded corners=3pt, minimum height=0.85cm, minimum width=1.5cm, align=center, font=\scriptsize\bfseries, fill=#1!12, line width=0.8pt},
        arrow/.style={-{Stealth[length=2mm]}, thick},
        label/.style={font=\scriptsize\itshape, text=gray!70!black},
    ]

    \node[block, fill=gray!8] (input) {Ab-Ag Complex\\$\{(s_i, \mathbf{X}_i)\}$};

    \node[block, fill=blue!8, above right=0.1cm and 0.8cm of input] (aa_emb) {AA Embed};
    \node[block, fill=blue!15, right=0.5cm of aa_emb, minimum width=1.8cm] (gnn) {RelationEGNN\\Encoder};
    \node[block, fill=orange!8, right=0.5cm of gnn, minimum width=1.3cm] (crop) {Crop\\Nearest AG};

    \node[block, fill=green!12, below right=0.1cm and 0.8cm of input, minimum width=2.0cm] (esm) {ESM-2\\(frozen/partial)};
    \node[block, fill=green!6, right=0.5cm of esm, minimum width=1.3cm] (esm_cdr) {CDR\\Embeddings};

    \path (input.center); \pgfgetlastxy{\inputx}{\inputy}
    \path (crop.east);    \pgfgetlastxy{\cropx}{\cropy}
    \node[stage=violet, minimum width=2.2cm, minimum height=1.4cm] (adapter) at (\cropx + 60pt, \inputy) {MiniStructural\\Adapter\\{\tiny CrossAttn + FFN}};

    \node[block, fill=blue!18, right=0.7cm of adapter, minimum width=1.5cm] (seq) {Sequence\\Head};

    \draw[arrow] (input.east) -- ++(0.2,0) |- (aa_emb.west);
    \draw[arrow] (input.east) -- ++(0.2,0) |- (esm.west);

    \draw[arrow] (aa_emb) -- (gnn);
    \draw[arrow] (gnn) -- (crop);
    \draw[arrow] (esm) -- (esm_cdr);

    \draw[arrow] (crop.east) -- (crop.east -| adapter.west) -- (adapter.north west) node[pos=0.0, right, font=\tiny] {KV};

    \draw[arrow] (esm_cdr.east) -- (esm_cdr.east -| adapter.west) -- (adapter.south west) node[pos=0.0, right, font=\tiny] {Q};

    \draw[arrow] (adapter) -- (seq);

    \node[label, below=0.15cm of gnn] {$\mathcal{L}_\text{coord}$};
    \node[label, below=0.15cm of seq] {$\mathcal{L}_\text{seq}$};

    \draw[decorate, decoration={brace, amplitude=4pt, raise=2pt}]
        (aa_emb.north west) -- (crop.north east) node[midway, above=6pt, font=\scriptsize\itshape] {Structural Encoder};
    \draw[decorate, decoration={brace, amplitude=4pt, mirror, raise=2pt}]
        (esm.south west) -- (esm_cdr.south east) node[midway, below=6pt, font=\scriptsize\itshape] {PLM Backbone };

    \end{tikzpicture}
    \caption{\textbf{\evostruct{} architecture.} Two parallel paths process the antibody-antigen complex. A Relational-EGNN encoder produces structural embeddings over the full complex graph, while frozen ESM-2 produces evolutionary embeddings for CDR positions. The Mini Structural Adapter cross-attends ESM-2 CDR queries to GNN-encoded structural keys/values (CDR + nearest antigen residues), whereas the Sequence Head produces the final CDR sequence designs. }
    \label{fig:architecture}
\end{figure*}

\evostruct{} processes the antibody-antigen complex through two parallel pathways and fuses their outputs via a cross-attention adapter (\cref{fig:architecture}), following the PLM-structure bridging paradigm of \citet{zheng2023lmdesign}. A relation-aware E(3)-equivariant GNN encodes the full 3D complex graph, capturing spatial geometry. Separately, a frozen protein language model produces CDR embeddings that encode evolutionary substitution patterns learned from hundreds of millions of protein sequences. A structural adapter then cross-attends the PLM's CDR embeddings to the GNN's structural context, and a sequence head operating in the PLM's representation space produces the final amino acid predictions. The entire sequence prediction pathway runs in the PLM's embedding space, preserving its calibrated amino acid vocabulary, while the GNN supplies structural context \emph{into} PLM representations.

\subsection{PLM Backbone}
\label{sec:esm}

The sequence pathway leverages ESM-2~\citep{lin2023evolutionary}, a protein language model trained on over 250 million protein sequences via masked language modeling. ESM-2 encodes amino acid substitution patterns, evolutionary conservation, and biochemical properties in its high-dimensional embeddings. We extract CDR representations by passing the full heavy-chain sequence through ESM-2 with CDR positions replaced by the mask token, then selecting the embeddings at CDR positions:
\begin{equation}
    \mathbf{H}_\text{esm} = \text{ESM-2}\!\left(\text{mask}(s_\text{HC}, V_\text{CDR})\right) \big|_{V_\text{CDR}} \in \mathbb{R}^{L \times d_\text{esm}}
    \label{eq:esm}
\end{equation}
where $L = |V_\text{CDR}|$ is the CDR length and $d_\text{esm}$ is the PLM embedding dimension. Masking CDR positions prevents trivial copying of the ground-truth sequence during training. The ESM-2 model remains frozen during early training and is progressively unfrozen in later phases following \citet{howard2018universal}, preventing catastrophic forgetting of evolutionary priors.

Operating in ESM-2's embedding space rather than the GNN's preserves the PLM's vocabulary calibration. ESM-2's output distribution, learned from hundreds of millions of sequences, naturally assigns non-zero probability to all 20 amino acids at each position, with probabilities reflecting evolutionary substitution patterns. GNN methods that learn their own amino acid embeddings from a few thousand training complexes lack this calibration and collapse to a handful of frequent amino acids under greedy decoding (\cref{sec:failure}). This is the same principle underlying LM-Design~\citep{zheng2023lmdesign} for general protein design, which we apply here to resolve the vocabulary collapse specific to CDR design.

\subsection{Structural Encoder}
\label{sec:encoder}

The structural pathway encodes the full antibody-antigen complex through multiple layers of a relation-aware E(3)-equivariant graph neural network~\citep{satorras2021n,wu2025raad} operating on graph $\mathcal{G}$ (\cref{sec:graph}). Each residue is initialized with a learned amino acid embedding, projected to the GNN hidden dimension via a linear layer. CDR residue embeddings are masked to zero during training. Epitope residues receive an additional learnable embedding that signals their membership in the designated binding site.

Each GNN layer updates node features and coordinates simultaneously. For edge $(i, j)$ of type $t$, the message function concatenates the sender and receiver embeddings, the outer product of coordinate displacements, and the edge features:
\begin{equation}
    \mathbf{m}_{ij}^{(l)} = \text{MLP}_\text{msg}^{(l)}\!\left([\mathbf{h}_i^{(l)},\; \mathbf{h}_j^{(l)},\; \text{vec}\!\left(\Delta\mathbf{x}_{ij} (\Delta\mathbf{x}_{ij})^\top\right),\; \mathbf{e}_{ij}]\right)
    \label{eq:message}
\end{equation}
where $\Delta\mathbf{x}_{ij} = \mathbf{x}_i^{(l)} - \mathbf{x}_j^{(l)}$ and $\text{vec}(\cdot)$ flattens the $3 \times 3$ outer product into a 9-dimensional vector. The outer product entries are dot products of displacement components, ensuring E(3)-invariance of the message function. Node features aggregate messages from all edge types with type-specific linear projections and a skip connection for training stability:
\begin{equation}
    \mathbf{h}_i^{(l+1)} = \mathbf{h}_i^{(l)} + \text{MLP}_\text{node}^{(l)}\!\left(\left[\mathbf{h}_i^{(l)},\; \sum_{t} \mathbf{W}_t^{(l)} \sum_{j \in \mathcal{N}_t(i)} \mathbf{m}_{ij}^{(l)}\right]\right)
    \label{eq:node_update}
\end{equation}
where $\mathbf{W}_t^{(l)}$ is a type-specific projection matrix and $\mathcal{N}_t(i)$ denotes the neighbors of node $i$ under edge type $t$. Coordinates are updated equivariantly by adding a weighted sum of displacement vectors:
\begin{equation}
    \mathbf{x}_i^{(l+1)} = \mathbf{x}_i^{(l)} + \sum_{t} \frac{1}{|\mathcal{N}_t(i)|}\sum_{j \in \mathcal{N}_t(i)} \Delta\mathbf{x}_{ij} \cdot \text{MLP}_t^{\text{coord},(l)}(\mathbf{m}_{ij}^{(l)})
    \label{eq:coord_update}
\end{equation}
The product $\Delta\mathbf{x}_{ij} \cdot \text{scalar}$ is equivariant by construction, because any rotation $\mathbf{R}$ transforms the displacement as $\mathbf{R}\Delta\mathbf{x}_{ij}$ while leaving the scalar factor unchanged. After $L_\text{enc}$ layers, the encoder produces per-residue embeddings $\mathbf{h} \in \mathbb{R}^{N \times d_\text{gnn}}$ and updated backbone coordinates $\hat{\mathbf{X}} \in \mathbb{R}^{N \times 4 \times 3}$.

\paragraph{Antigen context cropping.}
Rather than providing the full antigen to the adapter, we crop to the $K_\text{ag}$ antigen residues nearest to the CDR by C$\alpha$ distance. This bounds memory usage while preserving the most relevant epitope information, since binding contacts necessarily involve spatially proximal residues. The structural context passed to the adapter is the concatenation of CDR and cropped antigen embeddings:
\begin{equation}
    \mathbf{H}_\text{ctx} = [\mathbf{h}_k \mid k \in V_\text{CDR}] \,\|\, [\mathbf{h}_j \mid j \in \text{top}_{K_\text{ag}}\!(V_A)] \in \mathbb{R}^{(L + K_\text{ag}) \times d_\text{gnn}}
    \label{eq:ctx}
\end{equation}

\subsection{Mini Structural Adapter}
\label{sec:adapter}

The adapter bridges the two representation spaces by projecting ESM-2's CDR embeddings into a shared intermediate space, cross-attending to GNN-encoded structural context, and projecting back to the PLM's embedding dimension. This adapter contains a single cross-attention block followed by a feed-forward network (FFN), with residual connections and layer normalization at each stage.

We first project both representations into a shared adapter dimension $d_a$:
\begin{align}
    \mathbf{Q} &= \mathbf{H}_\text{esm} \, \mathbf{W}_\text{down} \in \mathbb{R}^{L \times d_a} \label{eq:proj_q} \\
    \mathbf{K}, \mathbf{V} &= \mathbf{H}_\text{ctx} \, \mathbf{W}_\text{gnn} \in \mathbb{R}^{(L + K_\text{ag}) \times d_a} \label{eq:proj_kv}
\end{align}
where $\mathbf{W}_\text{down} \in \mathbb{R}^{d_\text{esm} \times d_a}$ projects ESM-2 embeddings down and $\mathbf{W}_\text{gnn} \in \mathbb{R}^{d_\text{gnn} \times d_a}$ projects GNN embeddings up. The queries are derived from ESM-2 (evolutionary prior), while keys and values are derived from the GNN (structural context). This asymmetry encodes the design principle that the PLM representation queries ``what structural environment am I in?'' and the GNN provides the answer.

The multi-head cross-attention with $H$ heads computes attention scores and a weighted aggregation over the structural context:
\begin{equation}
    \text{CrossAttn}(\mathbf{Q}, \mathbf{K}, \mathbf{V}) = \Big\|_{h=1}^{H} \text{softmax}\!\left(\frac{\mathbf{Q}^{(h)} (\mathbf{K}^{(h)})^\top}{\sqrt{d_a / H}}\right) \mathbf{V}^{(h)}
    \label{eq:cross_attn}
\end{equation}
We then apply a residual connection and layer normalization after the cross-attention:
\begin{equation}
    \mathbf{Z} = \text{LayerNorm}\!\left(\mathbf{Q} + \text{CrossAttn}(\mathbf{Q}, \mathbf{K}, \mathbf{V})\right)
    \label{eq:attn_residual}
\end{equation}
A two-layer FFN with SiLU activation, also with residual connection and layer normalization, further processes the representations:
\begin{equation}
    \mathbf{Z}' = \text{LayerNorm}\!\left(\mathbf{Z} + \text{FFN}(\mathbf{Z})\right)
    \label{eq:ffn}
\end{equation}
Finally, the adapter projects back to the PLM's embedding dimension:
\begin{equation}
    \mathbf{H}_\text{refined} = \mathbf{Z}' \, \mathbf{W}_\text{up} \in \mathbb{R}^{L \times d_\text{esm}}
    \label{eq:proj_up}
\end{equation}
The refined representations inhabit ESM-2's embedding space but are now informed by the 3D structural context of the antibody-antigen complex, including the relative geometry of CDR and antigen residues and the edge-type-specific interactions encoded by the GNN.

\subsection{Sequence Head and Coordinate Prediction}
\label{sec:seq_head}

The sequence head maps the refined PLM-dimensional representations to amino acid logits through a two-layer MLP with layer normalization, SiLU activation, and dropout:
\begin{equation}
    \boldsymbol{\ell}_i = \text{MLP}_\text{seq}(\text{LayerNorm}(\mathbf{h}_i^\text{refined})) \in \mathbb{R}^{|\mathcal{A}|}
    \label{eq:logits}
\end{equation}
where $|\mathcal{A}|$ is the amino acid vocabulary size. At inference, we predict the amino acid at each position via greedy decoding, $\hat{s}_i = \argmax_{a} \ell_i^a$.

The GNN encoder simultaneously produces updated C$\alpha$ coordinates $\hat{\mathbf{x}}_k$ for CDR positions through the equivariant coordinate update (\cref{eq:coord_update}). These predicted coordinates provide the structural component of the co-design output.

\subsection{Training Objective}
\label{sec:objective}

We train \evostruct{} with five loss terms combined with an R-Drop consistency regularizer~\citep{liang2021rdrop}.

\paragraph{Sequence loss.}
We minimize the per-position cross-entropy on predicted CDR amino acids:
\begin{equation}
    \mathcal{L}_\text{seq} = -\frac{1}{L}\sum_{i=1}^{L} \log \frac{\exp(\ell_i^{y_i})}{\sum_{a} \exp(\ell_i^a)}
    \label{eq:ce}
\end{equation}
where $y_i$ is the ground-truth amino acid at CDR position $i$.

\paragraph{Coordinate loss.}
We apply a smooth-$\ell_1$ (Huber) loss on the GNN's predicted versus true C$\alpha$ coordinates for CDR positions:
\begin{equation}
    \mathcal{L}_\text{coord} = \frac{1}{L}\sum_{k \in V_\text{CDR}} \text{smooth}_{\ell_1}\!\left(\hat{\mathbf{x}}_k^{\text{C}\alpha} - \mathbf{x}_k^{\text{C}\alpha,\text{true}}\right)
\end{equation}

\paragraph{Pairing loss.}
We employ an InfoNCE-style contrastive loss~\citep{chen2020simple} that matches mean-pooled CDR and antigen GNN embeddings within the batch, encouraging the encoder to produce representations that distinguish cognate from non-cognate antibody-antigen pairs:
\begin{equation}
    \mathcal{L}_\text{pair} = -\frac{1}{B}\sum_{i=1}^{B} \log \frac{\exp(\bar{\mathbf{h}}_i^\text{cdr} \cdot \bar{\mathbf{h}}_i^\text{ag} / \tau_p)}{\sum_{k=1}^{B} \exp(\bar{\mathbf{h}}_i^\text{cdr} \cdot \bar{\mathbf{h}}_k^\text{ag} / \tau_p)}
\end{equation}
where $B$ is the batch size and $\tau_p$ is a temperature parameter.

\paragraph{Docking loss.}
We penalize the minimum predicted C$\alpha$ distance from each CDR residue to epitope atoms when it exceeds a cutoff, encouraging the predicted CDR backbone to dock near the epitope surface:
\begin{equation}
    \mathcal{L}_\text{dock} = \frac{1}{L}\sum_{k \in V_\text{CDR}} \max\!\left(0,\; \min_{j \in E} \|\hat{\mathbf{x}}_k - \mathbf{x}_j\| - d_\text{dock}\right)
\end{equation}

\paragraph{Shadow paratope loss.}
Following dyMEAN~\citep{kong2023dymean}, we penalize deviation of the predicted CDR-epitope pairwise distance matrix from ground truth. This provides a richer geometric supervisory signal than coordinate loss alone, because it supervises the relative arrangement of CDR and epitope residues rather than absolute positions:
\begin{equation}
    \mathcal{L}_\text{shadow} = \frac{1}{L|E|}\sum_{k \in V_\text{CDR}} \sum_{j \in E} \left|\|\hat{\mathbf{x}}_k - \mathbf{x}_j\| - \|\mathbf{x}_k^\text{true} - \mathbf{x}_j\|\right|
\end{equation}

\paragraph{R-Drop regularization.}
Each training step performs two forward passes with different dropout masks, producing two sequence losses $\mathcal{L}_\text{seq}^{(1)}$ and $\mathcal{L}_\text{seq}^{(2)}$. Inspired by R-Drop~\citep{liang2021rdrop}, we penalize inconsistency between the two passes. We use a squared loss difference rather than the symmetric KL divergence of the original formulation, as this avoids computing per-position KL terms while still encouraging consistent predictions:
\begin{equation}
    \mathcal{L}_\text{R\text{-}Drop} = \alpha_\text{rd} \cdot \left(\mathcal{L}_\text{seq}^{(1)} - \mathcal{L}_\text{seq}^{(2)}\right)^2
    \label{eq:rdrop}
\end{equation}
The consistency penalty encourages robust structural context injection regardless of which cross-attention connections are dropped. The total training loss averages the two base losses and adds the consistency penalty:
\begin{equation}
    \mathcal{L} = \frac{1}{2}\!\left(\mathcal{L}_\text{base}^{(1)} + \mathcal{L}_\text{base}^{(2)}\right) + \mathcal{L}_\text{R\text{-}Drop}
    \label{eq:total_loss}
\end{equation}
where $\mathcal{L}_\text{base} = \mathcal{L}_\text{seq} + \lambda_\text{coord}\mathcal{L}_\text{coord} + \lambda_\text{pair}\mathcal{L}_\text{pair} + \lambda_\text{dock}\mathcal{L}_\text{dock} + \lambda_\text{shadow}\mathcal{L}_\text{shadow}$.

\section{Experiments}
\label{sec:experiments}

\subsection{Experimental Setup}
\label{sec:setup}

\paragraph{Dataset and splits.}
We evaluate on the \chimera{} benchmark~\citep{ahmed2026chimerabench}, which contains 2,922 antibody-antigen complexes from the Structural Antibody Database (SAbDab) after deduplication at 95\% sequence identity. We use the epitope-group split, which clusters complexes by epitope similarity to test generalization to unseen binding sites.

\paragraph{Baselines.}
We compare against eleven CDR-H3 design methods spanning five architectural families. (1) \emph{GNN-based:} RAAD~\citep{wu2025raad}, MEAN~\citep{kong2022mean}, dyMEAN~\citep{kong2023dymean}. (2) \emph{Diffusion:} DiffAb~\citep{luo2022diffab}, AbFlowNet~\citep{abir2025abflownet}, AbMEGD~\citep{chen2025AbMEGD}. (3) \emph{Retrieval-augmented:} RADAb~\citep{wang2024radab}. (4) \emph{Flow/ODE:} dyAb~\citep{tan2025dyab}, AbODE~\citep{verma2023abode}. (5) \emph{Autoregressive:} RefineGNN~\citep{jin2021refinegnn}, AbDockGen~\citep{jin2022hern}. All baselines are retrained on \chimera{} using the authors' released code with default hyperparameters.

\paragraph{Metrics.}
We retrained the baseline methods on the \chimera{} dataset and evaluated based on multiple metrics. \emph{Sequence quality:} amino acid recovery (AAR), contact amino acid recovery (CAAR), and perplexity (PPL). \emph{Structural quality:} C$\alpha$ RMSD. \emph{Interface quality:} fraction of native contacts (fnat) and DockQ. \emph{Epitope awareness:} epitope F1. \emph{Safety:} number of sequence liabilities ($n_\text{liab}$).

\begin{figure}[t!]
    \centering
    \begin{tikzpicture}
    \begin{scope}[xscale=0.38, yscale=2.8]
        \draw[->] (0,0) -- (0,0.5) node[above, font=\tiny] {AAR};
        \foreach \y/\lab in {0.1/0.1, 0.2/0.2, 0.3/0.3, 0.4/0.4} {
            \draw (-0.15,\y) -- (0,\y) node[left, font=\tiny, xshift=-2pt] {\lab};
            \draw[gray!20] (0,\y) -- (12.5,\y);
        }

        \fill[red!60] (0.5,0) rectangle (1.3,0.43);
        \fill[blue!50] (1.8,0) rectangle (2.6,0.37);
        \fill[blue!40] (3.1,0) rectangle (3.9,0.37);
        \fill[blue!30] (4.4,0) rectangle (5.2,0.37);
        \fill[green!40] (5.7,0) rectangle (6.5,0.26);
        \fill[orange!40] (7.0,0) rectangle (7.8,0.24);
        \fill[orange!35] (8.3,0) rectangle (9.1,0.23);
        \fill[orange!30] (9.6,0) rectangle (10.4,0.23);
        \fill[purple!30] (10.9,0) rectangle (11.7,0.21);

        \node[font=\tiny, rotate=45, anchor=east] at (0.9,-0.03) {\textbf{EvoStruct.}};
        \node[font=\tiny, rotate=45, anchor=east] at (2.2,-0.03) {RAAD};
        \node[font=\tiny, rotate=45, anchor=east] at (3.5,-0.03) {MEAN};
        \node[font=\tiny, rotate=45, anchor=east] at (4.8,-0.03) {dyM.};
        \node[font=\tiny, rotate=45, anchor=east] at (6.1,-0.03) {AbODE};
        \node[font=\tiny, rotate=45, anchor=east] at (7.4,-0.03) {AbDG};
        \node[font=\tiny, rotate=45, anchor=east] at (8.7,-0.03) {DiffAb};
        \node[font=\tiny, rotate=45, anchor=east] at (10.0,-0.03) {AbFN};
        \node[font=\tiny, rotate=45, anchor=east] at (11.3,-0.03) {RefG.};

        \node[font=\tiny, red!80!black] at (0.9,0.45) {\textbf{.43}};
        \node[font=\tiny] at (2.2,0.39) {.37};
        \node[font=\tiny] at (3.5,0.39) {.37};
        \node[font=\tiny] at (4.8,0.39) {.37};
    \end{scope}
    \end{tikzpicture}
    \caption{\textbf{Amino acid recovery (AAR) comparison.} \evostruct{} (red) achieves 0.43 AAR, a 16\% relative improvement over the best GNN baselines (0.37). Blue: GNN methods. Orange: diffusion/flow. Green: ODE. Purple: autoregressive.}
    \label{fig:aar_bar}
\end{figure}
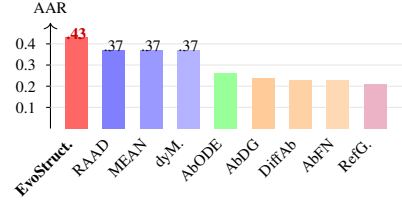

\begin{table*}[ht!]
    \caption{CDR-H3 design results on the \chimera{} benchmark. We report mean $\pm$ std across complexes. \textbf{Bold} indicates the best value, \uline{underline} indicates second best. ``--'' indicates the method does not produce perplexity scores.}
    \label{tab:main_results}
    \begin{center}
    \begin{small}
    \begin{tabular}{lcccccccc}
        \toprule
        Method & AAR $\uparrow$ & CAAR $\uparrow$ & PPL $\downarrow$ & RMSD $\downarrow$ & fnat $\uparrow$ & DockQ $\uparrow$ & EpiF1 $\uparrow$ & $n_\text{liab}$ $\downarrow$ \\
        \midrule
        \evostruct{} & \textbf{0.43\std{0.19}} & \second{0.22\std{0.27}} & \textbf{1.88\std{0.43}} & \second{1.84\std{0.82}} & \second{0.61\std{0.31}} & \second{0.70\std{0.19}} & \second{0.73\std{0.24}} & 0.70\std{0.84} \\
        \midrule
        RAAD & \second{0.37\std{0.12}} & 0.21\std{0.22} & 3.27\std{0.48} & \textbf{1.75\std{0.77}} & 0.56\std{0.30} & 0.70\std{0.15} & 0.72\std{0.25} & 0.57\std{0.65} \\
        MEAN & \second{0.37\std{0.13}} & \textbf{0.24\std{0.23}} & \second{3.10\std{0.47}} & 1.84\std{0.75} & 0.57\std{0.31} & 0.69\std{0.15} & 0.72\std{0.25} & 0.58\std{0.49} \\
        dyMEAN & \second{0.37\std{0.13}} & \second{0.22\std{0.23}} & 3.29\std{0.40} & 2.22\std{0.97} & 0.53\std{0.31} & 0.65\std{0.15} & 0.64\std{0.28} & 0.83\std{0.38} \\
        AbODE & 0.26\std{0.12} & 0.20\std{0.22} & 11.70\std{4.34} & 14.64\std{3.21} & 0.11\std{0.21} & 0.37\std{0.15} & 0.27\std{0.25} & 1.50\std{1.90} \\
        \midrule
        DiffAb & 0.23\std{0.12} & 0.14\std{0.19} & -- & 2.49\std{1.28} & 0.59\std{0.31} & 0.65\std{0.16} & 0.64\std{0.25} & 0.52\std{0.71} \\
        AbFlowNet & 0.23\std{0.11} & 0.14\std{0.18} & -- & 2.38\std{1.22} & 0.60\std{0.31} & 0.66\std{0.16} & 0.65\std{0.25} & \second{0.38\std{0.63}} \\
        AbMEGD & 0.21\std{0.12} & 0.12\std{0.16} & -- & 2.44\std{1.29} & 0.56\std{0.29} & 0.64\std{0.14} & 0.64\std{0.25} & 0.49\std{0.67} \\
        RADAb & 0.20\std{0.12} & 0.11\std{0.17} & -- & 5.33\std{17.22} & 0.49\std{0.32} & 0.60\std{0.17} & 0.60\std{0.27} & 0.56\std{0.74} \\
        dyAb & 0.19\std{0.08} & 0.09\std{0.14} & -- & 2.34\std{0.87} & 0.14\std{0.21} & 0.45\std{0.09} & 0.24\std{0.31} & \textbf{0.00\std{0.00}} \\
        \midrule
        RefineGNN & 0.21\std{0.11} & 0.10\std{0.14} & 8.46\std{3.28} & 2.86\std{0.87} & \textbf{0.65\std{0.28}} & \textbf{0.73\std{0.14}} & \textbf{0.76\std{0.22}} & 0.71\std{0.80} \\
        AbDockGen & 0.24\std{0.12} & 0.12\std{0.18} & 8.04\std{2.65} & 4.67\std{1.32} & 0.41\std{0.26} & 0.55\std{0.14} & 0.62\std{0.22} & 0.70\std{0.79} \\
        \bottomrule
    \end{tabular}
    \end{small}
    \end{center}
\end{table*}

\subsection{Implementation Details}
\label{sec:impl}

We use ESM-2 with 650M parameters ($d_\text{esm} = 1280$) as the PLM backbone. The RelationEGNN encoder has 5 layers with hidden dimension $d_\text{gnn} = 256$, amino acid embedding dimension 32, and 8 edge types. We crop the $K_\text{ag} = 128$ nearest antigen residues as structural context. The adapter operates in dimension $d_a = 640$ with $H = 8$ attention heads and an FFN expansion ratio of 2. The sequence head projects from 1280 to 640 to 25 dimensions. 


We train with Adam optimizer, gradient clipping at 0.5, dropout rate 0.2, and batch size 4, using early stopping with patience 10 on validation loss. The loss weights are determined by hyperparameter sweep ($\lambda_\text{coord} = 1.376$, $\lambda_\text{pair} = 0.525$, $\lambda_\text{dock} = 0.5$, $\lambda_\text{shadow} = 0.3$, $\alpha_\text{rd} = 1.0$). Training follows a three-phase progressive unfreezing schedule~\citep{howard2018universal}. Phase 1 (up to 50 epochs, lr $= 10^{-4}$) trains the adapter, GNN, and sequence head with ESM-2 frozen. Phase 2 (40 epochs, lr $= 5 \times 10^{-5}$) unfreezes the top 4 ESM-2 transformer layers. Phase 3 (30 epochs, lr $= 10^{-5}$) fine-tunes all parameters at a low learning rate. All phases use exponential LR decay ($\gamma = 0.9$). It takes approximately 3 hours to train the model on a single H100 GPU.

\subsection{Main Results}
\label{sec:results}

\begin{figure*}[h!]
    \includegraphics[width=\textwidth]{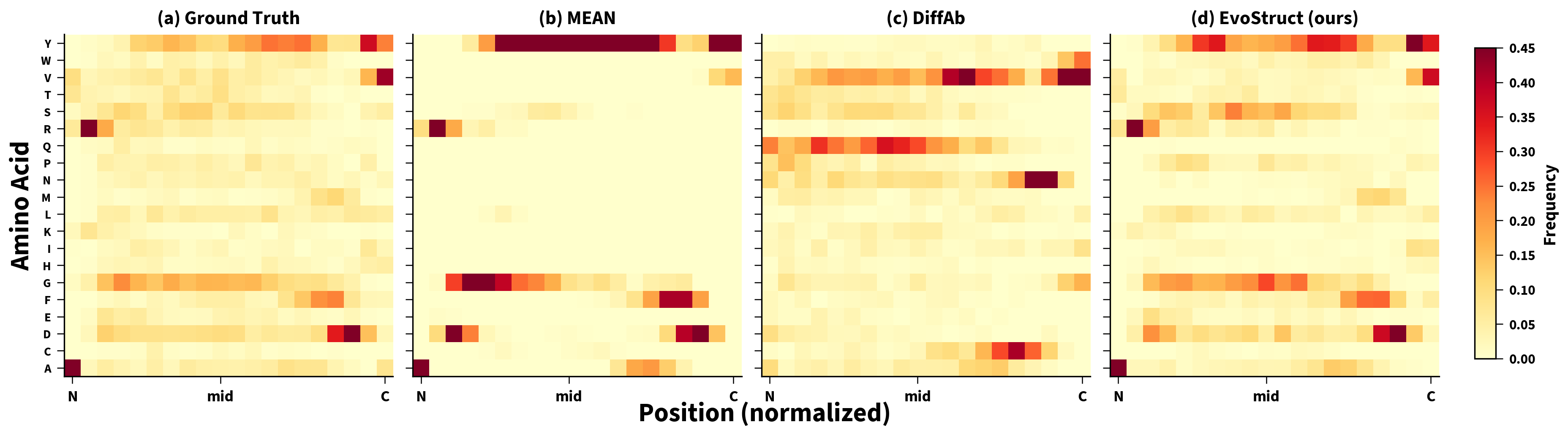}
        \caption{\textbf{Vocabulary collapse.} Per-position amino acid frequency heatmaps. \evostruct{} preserves near-native diversity ($V_\text{eff}{=}12.4$).}
        \label{fig:vocab_collapse}
\end{figure*}

\Cref{tab:main_results} presents the main comparison. \evostruct{} achieves the highest amino acid recovery (0.43) among all eleven methods, a 16\% relative improvement over the best GNN baselines (RAAD, MEAN, dyMEAN at 0.37). The improvement is accompanied by a 43\% reduction in perplexity (1.88 vs. 3.27 for RAAD), indicating substantially more confident and calibrated predictions. Structurally, \evostruct{} produces CDR backbones with RMSD 1.84~\AA, comparable to RAAD (1.75~\AA) and MEAN (1.84~\AA). Interface metrics (fnat 0.61, DockQ 0.70) are competitive with the best GNN baselines, and epitope F1 (0.73) matches the GNN range.

RefineGNN achieves the highest fnat (0.65) and DockQ (0.73) despite not conditioning on the antigen, confirming the finding from prior analyses~\citep{ahmed2026chimerabench} that antigen conditioning does not currently improve binding quality in existing architectures. \evostruct{} narrows this gap while simultaneously leading in sequence recovery, suggesting that PLM-derived representations offer a path toward methods that excel in both sequence and structural quality.

\begin{figure*}[ht!]
        \includegraphics[width=\textwidth]{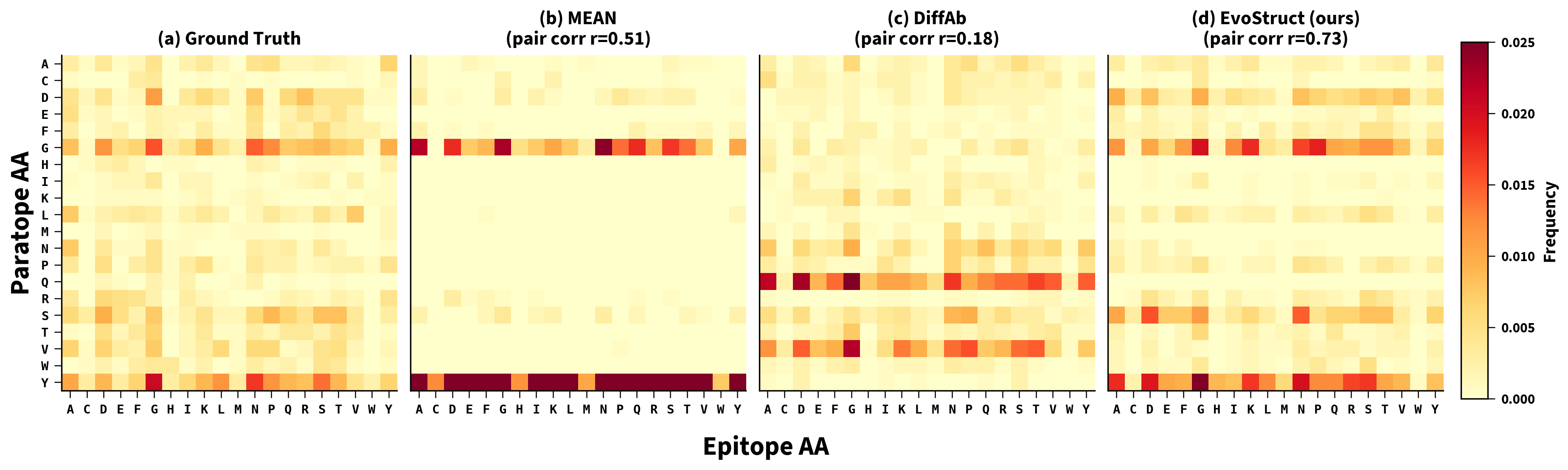}
        \caption{\textbf{Binding-pair preferences.} Paratope-epitope AA pair heatmaps. \evostruct{} achieves the highest pair correlation ($r{=}0.73$) with ground truth, capturing binding-specific preferences.}
        \label{fig:binding_pairs}
\end{figure*}

\subsection{Failure Mode Resolution}
\label{sec:failure_resolution}

We evaluate \evostruct{} on the three failure modes identified in \cref{sec:failure}.

\begin{figure*}[h!]
        \includegraphics[width=\textwidth]{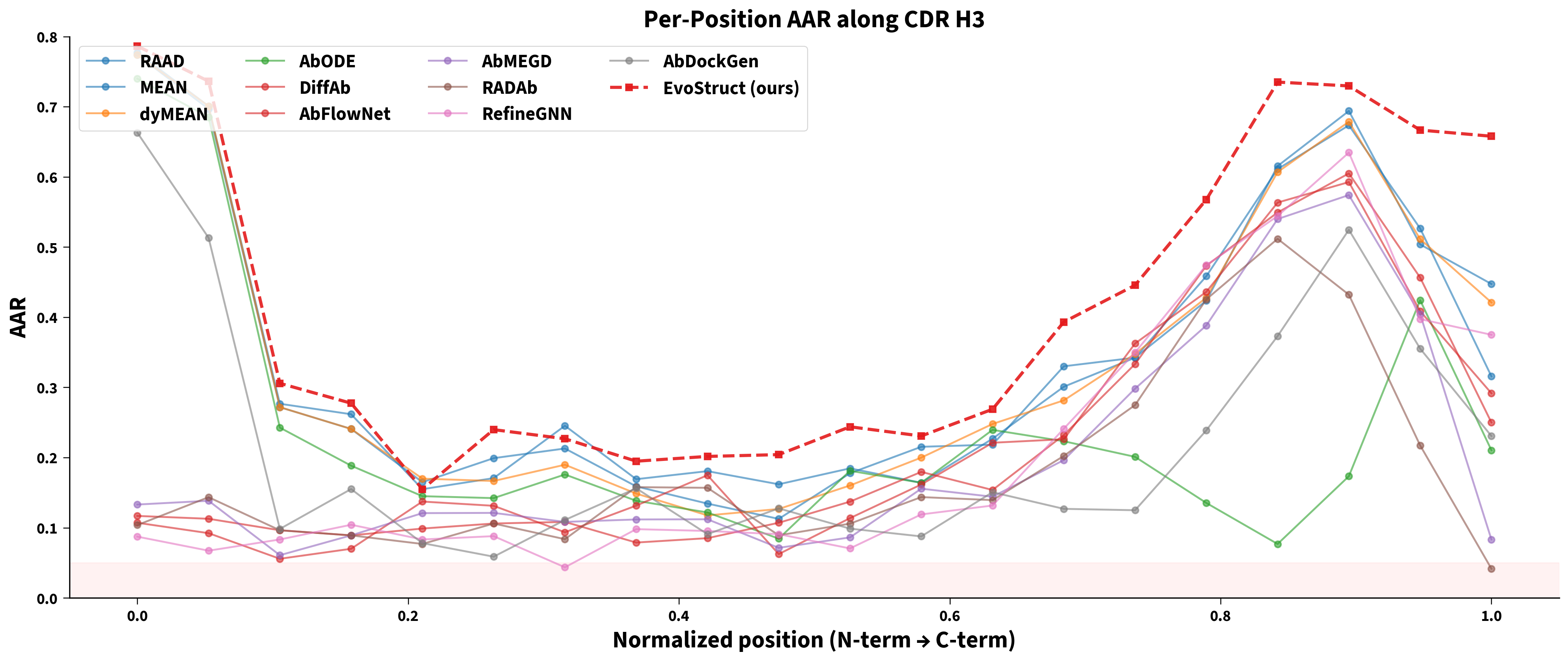}
        \caption{\textbf{Positional AAR profile.} Per-position recovery along CDR-H3. \evostruct{} (red dashed) leads at both anchor positions and the hypervariable apex.}
        \label{fig:positional}
\end{figure*}

\paragraph{Vocabulary diversity.}
\evostruct{} achieves $V_\text{eff} = 12.4$, recovering 80\% of the ground-truth amino acid diversity (\cref{fig:vocab_collapse}). By contrast, the best GNN baselines (RAAD, MEAN, dyMEAN) collapse to $V_\text{eff}$ of 4.7--5.3, recovering only 30--34\% of native diversity. \evostruct{} produces 282 unique bigrams and 1,214 unique trigrams (vs. 364/1,818 in the ground truth and 52/110 for RAAD), covering all 20 of the top ground-truth motifs. This demonstrates that operating in ESM-2's embedding space preserves the PLM's calibrated vocabulary, breaking the diversity-accuracy tradeoff that constrains GNN-only methods.

\paragraph{Binding-pair correlation.}
\evostruct{} achieves the highest interface enrichment correlation with ground truth ($r = 0.81$ vs. next-best 0.68 for RAAD) and the highest paratope-epitope pair correlation ($r = 0.73$ vs. next-best 0.69 for AbFlowNet), as shown in \cref{fig:binding_pairs}. This indicates that the structural adapter successfully routes antigen information to the sequence head, enabling \evostruct{} to learn which amino acids pair with which epitope residues.

\paragraph{Positional recovery.}
\evostruct{} improves AAR at both contact positions (22.6\% vs. 20.6\% for RAAD) and non-contact positions (51.0\% vs. 43.5\% for RAAD), as shown in \cref{fig:positional}. The larger gain at non-contact positions (+7.5pp) reflects ESM-2's strong framework priors, which provide better predictions at conserved positions not constrained by the binding partner. The per-position AAR profile along CDR-H3 shows that \evostruct{} dominates at anchor positions (71.2\% vs. 60.7\% for RAAD), where ESM-2's evolutionary priors are strongest, and improves at the hypervariable apex (22.4\% vs. 19.0\%). Contact AAR remains the hardest challenge for all methods, suggesting that deeper antigen conditioning mechanisms beyond cross-attention may be needed.


\subsection{Discussion}
\label{sec:discussion}

The results confirm that the PLM-structure adapter paradigm~\citep{zheng2023lmdesign}, originally developed for general protein inverse folding, transfers effectively to CDR design. The vocabulary calibration of ESM-2 transfers directly through the adapter, without requiring the model to relearn amino acid substitution patterns from limited structural data. GNN methods must discover these patterns from ${\sim}$3,000 training complexes, which leads them to converge on the few most frequent amino acids per position. \evostruct{} avoids this by inheriting ESM-2's distribution over all 20 amino acids, achieving higher diversity and higher accuracy simultaneously.

The adapter architecture enables a clear separation of concerns. The GNN encoder focuses on encoding 3D geometry and spatial relationships, while ESM-2 handles amino acid vocabulary calibration and evolutionary substitution patterns. The cross-attention mechanism allows structural context to modulate evolutionary priors without overwriting them.

The main limitation is that \evostruct{} does not achieve the highest binding metrics. RefineGNN, which does not condition on the antigen at all, achieves fnat 0.65 and DockQ 0.73 compared to \evostruct{}'s 0.61 and 0.70. This suggests that the cross-attention adapter, while effective for sequence quality, does not fully exploit antigen structural information for binding interface prediction. Contact AAR remains at 22.6\% across methods, pointing to a fundamental difficulty in predicting antigen-specific amino acid identity at binding positions that may require more expressive conditioning mechanisms or alternative training objectives. Future work could explore richer adapter architectures, alternative sequence objectives that encourage antigen-specific predictions at contact positions, and integration with explicit contact prediction mechanisms.

\section{Conclusion}
\label{sec:conclusion}

In this paper, we diagnose vocabulary collapse as a systematic failure mode in GNN-based CDR design and resolve it by adapting the PLM-structure bridging paradigm for CDR design. \evostruct{} cross-attends ESM-2 CDR embeddings to GNN-encoded structural features, keeping the sequence prediction pathway in the PLM's calibrated representation space. Several experiments on the \chimera{} benchmark demonstrate that \evostruct{} breaks the diversity-accuracy tradeoff, achieving the highest sequence recovery and vocabulary diversity among eleven baselines with the strongest binding-pair correlation.



\section*{Impact Statement}

This paper presents work whose goal is to advance computational antibody design. The designed sequences require extensive experimental validation before any therapeutic application. We see no specific negative societal consequences that must be highlighted.

\newpage
\bibliography{references}
\bibliographystyle{icml2026}

\newpage

\section{Appendix}

\subsection{Detailed Results}

\begin{figure*}[h!]
    \includegraphics[width=\textwidth]{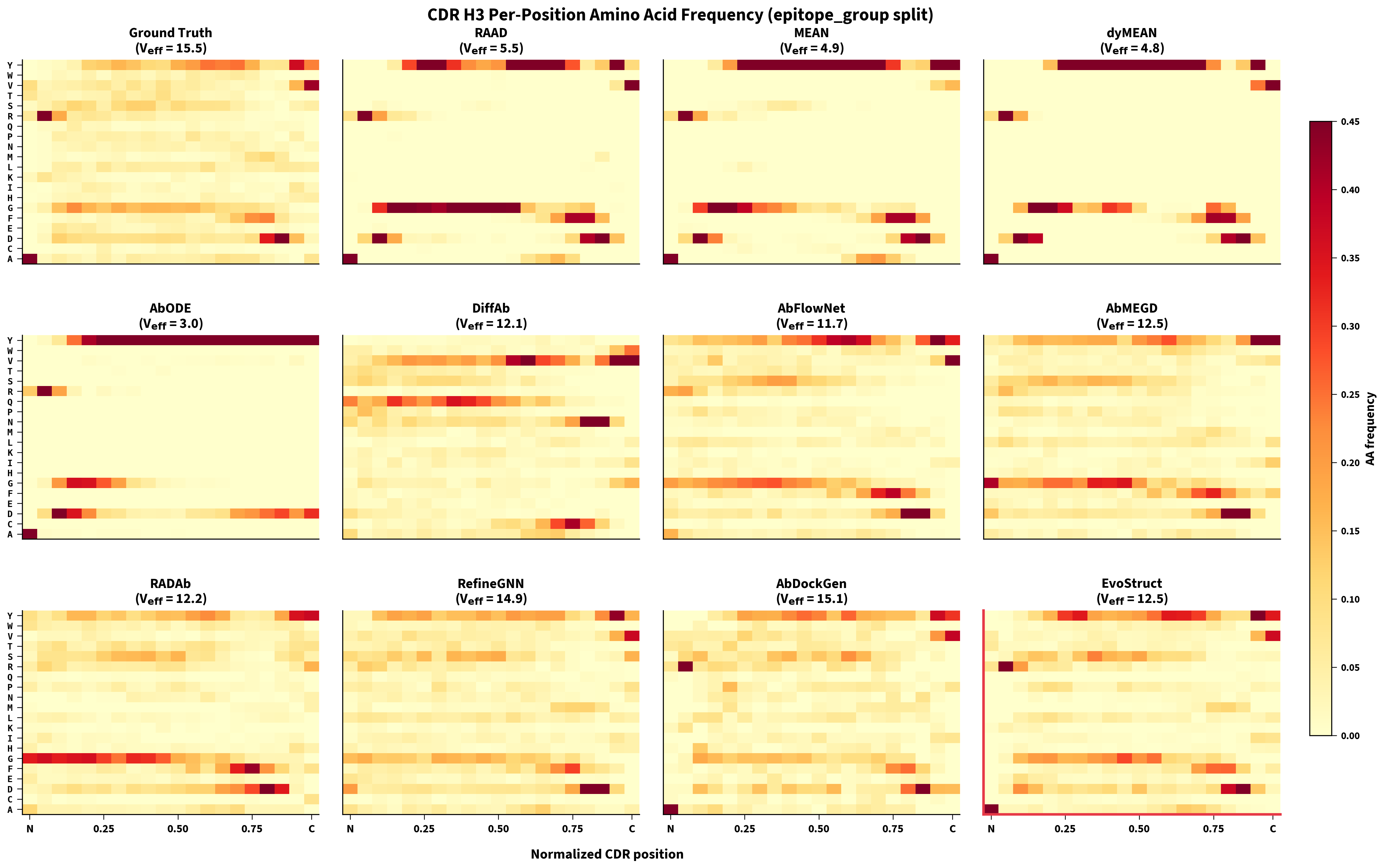}
        \caption{\textbf{Vocabulary collapse.} Per-position amino acid frequency heatmaps across all the benchmarked baseline methods.}
        \label{fig:vocab_collapse-all}
\end{figure*}

\begin{figure*}[h!]
        \includegraphics[width=\textwidth]{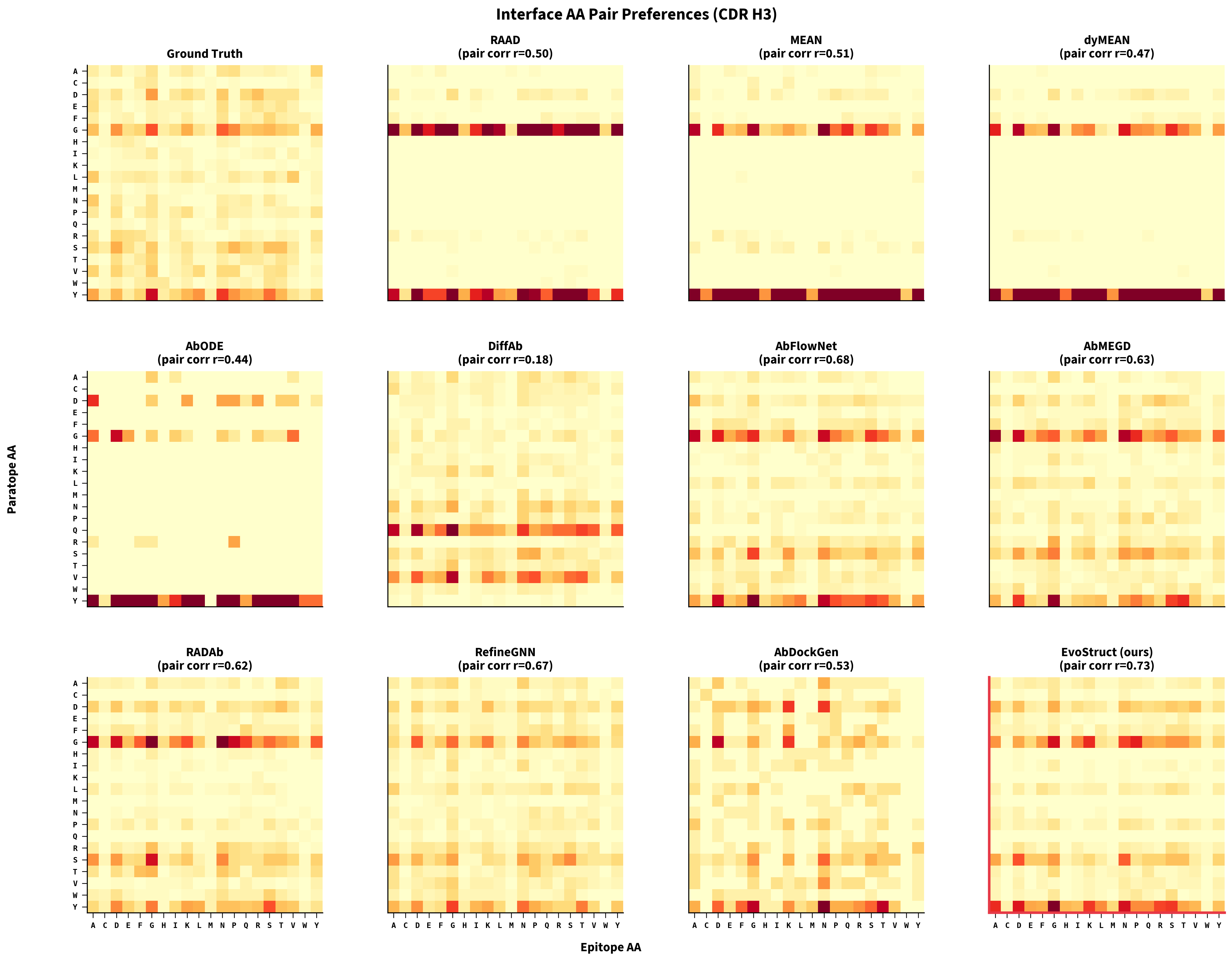}
        \caption{\textbf{Binding-pair preferences.} Paratope-epitope AA pair heatmaps across all the benchmarked baseline methods.}
        \label{fig:binding_pairs-all}
\end{figure*}

Below, we provide the detailed results for our benchmarking of the baseline methods and compare them with \evostruct{}. \cref{fig:vocab_collapse-all} shows the vocabulary collapse failure mode, while \cref{fig:binding_pairs-all} shows the failures of the baselines with regard to their binding pairs preference.

\end{document}